\documentclass{article}

\PassOptionsToPackage{numbers, compress}{natbib}

\usepackage[final]{bdl_2018}

\usepackage{graphicx}
\usepackage{amsmath}
\usepackage{mathtools}
\usepackage[utf8]{inputenc} 
\usepackage[T1]{fontenc}    
\usepackage{hyperref}       
\usepackage{url}            
\usepackage{booktabs}       
\usepackage{amsfonts}       
\usepackage{nicefrac}       
\usepackage{microtype}      

\title{Calibrating Uncertainties in Object Localization Task}

\author{
    Buu~Phan\footnotemark[1] \\
    \texttt{btphan@uwaterloo.ca} \\
    \And
    Rick~Salay\footnotemark[1] \\
    \texttt{rsalay@gsd.uwaterloo.ca} \\
    \And
    Krzysztof~Czarnecki\thanks{Department of Electrical and Computer Engineering, University of Waterloo} \\
    \texttt{kczarnec@gsd.uwaterloo.ca} \\
    \And
    Vahdat~Abdelzad\footnotemark[1] \\
    \texttt{vabdelza@uwaterloo.ca} \\
    \And
    Taylor~Denouden\thanks{Department of Computer Science, University of Waterloo} \\
    \texttt{tadenoud@uwaterloo.ca} \\
    \And
    Sachin~Vernekar\footnotemark[2] \\
    \texttt{sverneka@uwaterloo.ca} \\
}

\begin{document}

\maketitle

\begin{abstract}
  In many safety-critical applications such as autonomous driving and surgical robots, it is desirable to obtain prediction uncertainties from  object detection modules to help support safe decision-making. Specifically, such modules need to estimate the probability of each predicted object in a given region and the confidence interval for its bounding box. While recent Bayesian deep learning methods provide a principled way to estimate this uncertainty, the estimates for the bounding boxes obtained using these methods are uncalibrated. In this paper, we address this problem for the single-object localization task by adapting an existing technique for calibrating regression models. We show, experimentally, that the resulting calibrated model obtains more reliable uncertainty estimates.
\end{abstract}

\section{Introduction}

In safety-critical systems such as self-driving cars, it is desirable to have an object detection system that provides accurate predictions and reliable, or well-calibrated, associated uncertainties. The uncertainties in this case come from two sources: the bounding box regressor and the object classifier. For the bounding box regressor, a $p$\% confidence interval for each coordinate (estimated from the calibrated uncertainties) should contain the true value $p$\% of the time. Similarly, in classification, calibration means that predictions with $p$\% confidence are accurate $p$\% of the time. Miscalibration, in either case, can lead to dangerous situations in autonomous driving. For instance, if the detection model indicates the 95\% confidence interval that the location of a pedestrian is within the sidewalk, but it is actually a 50\% confidence interval, then the vehicle may make hazardous movements.   

Recent advances in Bayesian neural networks (BNNs) (\citet{gal2016dropout}, \citet{pmlr-v80-khan18a}, \citet{NIPS2017_7141}) have provided a framework for estimating uncertainties in deep neural networks (DNNs). The obtained uncertainty estimates from BNNs, however, requires calibration (\citet{pmlr-v80-kuleshov18a}, \citet{gal2016dropout}).

Recent works by \citet{miller2017dropout} and \citet{feng2018towards} have shown the benefits of modeling uncertainty for detection accuracy in 2D open-set conditions and the 3D Lidar object detection task, respectively. However, neither of them has analyzed or focused on the reliability of the estimated localization uncertainties in terms of calibration. In this paper, we address this issue for the 2D single object classification and localization (SOCL) task and demonstrate its applicability on the Oxford-IIIT Pet Dataset (\citet{parkhi12a}). We focus on the localization uncertainty estimates since our experiment shows that while BNNs produce a calibrated classification uncertainty (similar to \citet{mcclure2016representing}'s results), the estimated localization uncertainty is not calibrated. Specifically, our contributions are: (1) we show that the estimated localization uncertainties for the bounding box coordinates from the BNN model are not calibrated; (2) we adapt  \citet{pmlr-v80-kuleshov18a}'s method for calibrating regression models and show improvements in this setting. 

The remainder of the paper is structured as follows. Section 2 gives the necessary background for the SOCL task. Section 3 describes the calibration method for localization (see \citet{guo2017calibration} for calibration in classification). Section 4 shows experimental results demonstrating the method is effective. Finally, in Section 5 we discuss conclusions and future work. 
\section{Background: SOCL Task with Uncertainty Estimation}\label{2.1}
The goal of the SOCL task is to obtain a model that is able to predict the bounding box and class of an object in a given image. For an image dataset $\mathbf{X} = \{\mathbf{x}_1, \mathbf{x}_2, ..., \mathbf{x}_N\}$, the associated labels are $\mathbf{Y} = \{\mathbf{y}_1, \mathbf{y}_2, ..., \mathbf{y}_N\}$, where each $\mathbf{y}_k = [\mathbf{c}_k,\mathbf{b}_k]$ consists of a one-hot encoded class  $\mathbf{c}_k$ and the bounding box coordinates $\mathbf{b}_k = \{b_{1k},b_{2k},b_{3k},b_{4k}\}$. We define a DNN $\mathbf{f}_{\mathbf{W}}(\mathbf{x}) = \Hat{\mathbf{y}}$ with weights $\mathbf{W}$ such that it predicts both the class probability and coordinates.
\paragraph{Incorporating Uncertainty}
Based on \citet{NIPS2017_7141}'s method, we incorporate the aleatoric and epistemic uncertainties into the model by optimizing the weights $\mathbf{W}$ with heteroscedastic loss and dropout training (\citet{srivastava2014dropout}). At test time, we sample the predictions with MC-dropout and calculate the predictive mean and uncertainties. This results in a BNN model $\Bar{\mathbf{f}}_{\mathbf{W}}(\mathbf{x}) = [\Bar{\mathbf{y}}, \Bar{\boldsymbol{\sigma}}^2]$, where $\Bar{\mathbf{y}} = [\Bar{\mathbf{c}}, \Bar{\mathbf{b}}]$ is the predictive mean and $\Bar{\boldsymbol{\sigma}}^2 = \{\Bar{\sigma}^2_1,\Bar{\sigma}^2_{2},\Bar{\sigma}^2_{3},\Bar{\sigma}^2_{4}\}$ is the predictive variance (sum of the epistemic and aleatoric variance) of the coordinates, assuming that they are mutually independent \footnote[1]{This assumption gives an overapproximation of the bounding box extents, which is sufficient for obstacle avoidance.}. Similarly to \citet{lakshminarayanan2017simple}, we estimate the probability  $p(b_{i}|\mathbf{x})$ as a Gaussian: $p(b_{i}|\mathbf{x}) = \mathcal{N}(\Bar{b}_i, \Bar{\sigma}^2_i), \text{for }i\in {1,2,3,4}$.

\section{Calibration Method for Estimated Localization Uncertainty} \label{2.2}

We use the cumulative distribution function (CDF) form of $p(b_{i}|\mathbf{x})$ in Section \ref{2.1} for calibration, denoted as $P_{b_{i}|\mathbf{x}}(z) = \Phi(z|\Bar{b}_i, \Bar{\sigma}^2_i)$ where $\Phi(z|\mu,\gamma^2)$ is the CDF of Gaussian distribution $\mathcal{N}(\mu,\gamma^2)$ with mean $\mu$ and variance $\gamma^2$. For real value $z\in \mathbb{R}$,   $P_{b_{i}|\mathbf{x}}(z)$  is the probability that the label $b_i$ is in the $(-\infty, z]$ interval. Conversely, for a probability value $q$, we obtain the output $z$ of the inverse CDF: $P^{-1}_{b_{i}|\mathbf{x}}(q) = z$, which means that the interval $(-\infty, z]$ is a $100q$\% interval for $b_i$.

Let $\mathbb{I}_{b_{i}|\mathbf{x}}(q) \coloneqq \mathbb{I}[b_{i} \leq P^{-1}_{b_{i}|\mathbf{x}}(q)]$ be an indicator function that verifies the condition in the brackets. Then $P_{b_{i}|\mathbf{x}}(z)$ is calibrated when: 
\begin{equation}\label{eq1}
    \mathbb{E}[\mathbb{I}_{b_{i}|\mathbf{x}}(q)] = q
\end{equation}
This implies that we expect to see the $100q$\% confidence interval to cover $100q$\% of the label data. Calibrating a regression model means that we adjust $P_{b_{i}|\mathbf{x}}(z)$ such that (\ref{eq1}) holds. 
To obtain a reliable uncertainty estimate, we carry out two steps: validating the uncertainty estimate and calibrating it.

\paragraph{Validating the Uncertainty Estimates} In regression, higher estimated variance should correspond to higher expected square error. However, in practice, if the model lacks of expressiveness or does not converge, the resulting uncertainty estimates may not be valid and the calibration process will not give desired results. Thus, we use scatter plot (representing $\Bar{\sigma}^2_i$ and $(b_i - \Bar{b}_i)^2$) as a visualization method to validate this attribute of the uncertainty estimate before calibrating it. 

\paragraph{Calibrating the SOCL BNN} We adapt \citet{pmlr-v80-kuleshov18a} method of calibrating a BNN-based regressor for the case of bounding box estimation. Given an uncalibrated probabilistic model $\Bar{\mathbf{f}}_{\mathbf{W}}(\mathbf{x})$ for the SOCL task with $P_{b_{i}|\mathbf{x}}(z) = \Phi(z|\Bar{b}_i, \Bar{\sigma}^2_i)$ for each coordinate and a calibration dataset $\Hat{\mathbf{X}}$, $\Hat{\mathbf{Y}}$, the calibration process trains a calibration model $\emph{R}_i$ whose input is ${P}_{b_{i}|\mathbf{x}}(z)$  such that $\Hat{P}_{b_{i}|\mathbf{x}}(z) = \emph{R}_i \circ {P}_{b_{i}|\mathbf{x}}(z) = \emph{R}_i \circ \Phi(z|\Bar{b}_i, \Bar{\sigma}^2_i)$ is calibrated (see [8] for more details). After obtaining $R_i$, we replace ${P}_{b_{i}|\mathbf{x}}(z)$ by $\Hat{P}_{b_{i}|\mathbf{x}}(z)$ for localization uncertainty estimation.

We use $[x_{min},y_{min},x_{max},y_{max}]$ for encoding the coordinates in our experiments, where $x_{min}$,$y_{min}$ is the top left and $x_{max}$, $y_{max}$ is the bottom right corner of the bounding box.
To estimate the $100q$\% confidence interval around the mean, we determine the upper bound and lower bound for each $b_i$ by calculating $\Hat{P}^{-1}_{b_{i}|\mathbf{x}}(r + q/2)$ and $\Hat{P}^{-1}_{b_{i}|\mathbf{x}}(r - q/2)$ accordingly for each coordinate, where $r =\Hat{P}_{b_{i}|\mathbf{x}}(\Bar{b}_i) $. These bounds define a confidence interval as a region in which the bounding box can occur (see blue region in Figure. \ref{fig1}d).

\section{Experiments}\label{sec4}
In the following setting, we experimentally show the miscalibration problem of localization uncertainty of our model and the improvement after applying the calibration method on the model's output. For the completeness of the task, we also show the result for classification.

We use the Oxford-IIIT Pet Dataset (\citet{parkhi12a}), which consists of 3,686 annotated images in 2 classes depicting cats and dogs, for this task. The bounding box for localization covers the face of the pet. 
The dataset is split into 2:0.9:0.9 ratio for training, validation/calibration and testing respectively. 
For the model, the VGG-16 architecture (\citet{simonyan2014very}) is used as a base network. The trained model obtained 94.85\% classification accuracy and 14.03\% localization error with 0.5 IOU threshold, based on the Imagenet evaluation method (\citet{imagenet_cvpr09}).We validated the uncertainty estimates (epistemic and aleatoric) and fit the calibration model $R_i$ for each coordinate as described in Section 3.
\begin{figure}
\hspace*{-0.9cm}
  \centering
  \includegraphics[width=16cm]{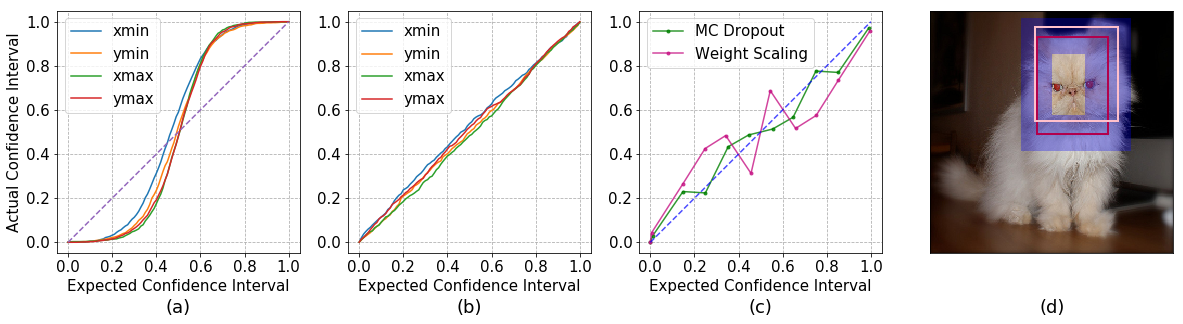}
  \caption{Reliability diagram and an example. Figure 1(a): reliability diagram for localization uncertainties before calibration. Figure 1(b): reliability diagram for localization uncertainties after calibration. Figure 1(c): reliability diagram for classification with MC-dropout and weight scaling. Figure 1(d) shows an example of bounding box localization with calibrated 95\% confidence interval (blue region) centered around the mean (red). The ground truth is colored in pink}
  \label{fig1}
\end{figure}
\paragraph{Results} Figure.\ref{fig1}a -\ref{fig1}c show the reliability diagram for localization and classification uncertainties on the test set. The reliability diagram shows the mapping between the expected confidence interval (from the model) and the actual one (i.e., how many labels are actually within that interval). Perfect calibration corresponds to the diagonal line in the diagram. Quantitatively, the calibration quality is evaluated by using the mean squared error (MSE) between the diagonal line and the calibration curve. Models with lower MSE are better calibrated.

Figure.\ref{fig1}a and \ref{fig1}b show the reliability diagram for bounding box coordinates before and after calibration respectively. Consider the curve for $x_{max}$ in Figure.\ref{fig1}a, we can see that the expected 40\% confidence interval corresponds to the 20\% actual interval. In this case, the model has underestimated the uncertainty. On the other hand, the expected 60\% confidence interval corresponds to the 80\% actual interval, which implies that the model has overestimated the uncertainty in this range. After calibration, the estimated uncertainties are reliable, e.g, the expected 20\% confidence interval contains approximately 20\% of the true outcome. The calibration process reduces the average MSE from 2.7E-02 to 2.7E-04 for the four coordinates.

We also observed that, in Figure.\ref{fig1}c, MC-dropout produces a calibrated classification confidence with MSE of 3.0E-03, compared to that of 1.6E-02 with the original weight scaling method for network trained with dropout (\citet{srivastava2014dropout}), in which we scale the weights according to the dropout rate at test time, and show the resulting softmax probabilities.


\section{Conclusion and Future Work}
In this paper, we consider the reliability of estimated localization uncertainty for the 2D SOCL task in terms of calibration. Our experiment shows that without calibration, the estimated localization uncertainties are misleading. We adapted an existing method for calibrating regression to the uncertainty estimates of bounding box coordinates. The result shows that the new uncertainty estimates are well-calibrated. 

In future work, we would like to extend this work to the general 2D and 3D multiple object detection task and use a more complex dataset such as KITTI (\citet{geiger2013vision}). Furthermore, we want to address the calibration problem in the case the coordinates are not mutually independent. Finally, although the computation cost for calibration step and estimating the aleatoric uncertainty is negligible, the cost for estimating the epistemic uncertainty is high. To reduce the computation
cost for this step, we want to investigate how model compression technique (\citet{han2015deep_compression}) can be incorporated for speeding up the inference rate. 
\bibliographystyle{plainnat}
\bibliography{refs.bib}
\end{document}